\pdfoutput=1

\documentclass[11pt]{article}

\usepackage[]{ACL2023}
\usepackage{times}
\usepackage{latexsym}

\usepackage[T1]{fontenc}

\usepackage[utf8]{inputenc}

\usepackage{microtype}

\usepackage{inconsolata}
\usepackage{listings}
\usepackage{comment}
\usepackage[colorinlistoftodos]{todonotes}
\usepackage{lineno}
\usepackage{tabularx}
\usepackage{booktabs}
\usepackage{mathtools}
\usepackage{pdflscape}
\usepackage{afterpage}
\usepackage{capt-of}
\usepackage{rotating}
\usepackage{colortbl}
\usepackage{soul}
\usepackage{xcolor}
\definecolor{Yellow}{RGB}{254,221,0}

\newsavebox{\fmbox}
\newenvironment{fmpage}[1]
{\begin{lrbox}{\fmbox}\begin{minipage}{#1}}
{\end{minipage}\end{lrbox}\fbox{\usebox{\fmbox}}} 

\usepackage{amsmath}

\title{Exploring Automatic Text Simplification of German Narrative Documents}

\author{Thorben Schomacker\\
  Hamburg Univ. of Applied Sciences\\
  \texttt{thorben.schomacker@}\\\texttt{haw-hamburg.de} 
  \\\And
  Tillmann Dönicke\\
  Univ. of Göttingen\\
  \texttt{tillmann.doenicke@}\\\texttt{uni-goettingen.de}
  \\\And
  Marina Tropmann-Frick\\
  Hamburg Univ. of Applied Sciences\\ 
  \texttt{marina.tropmann-frick@}\\\texttt{haw-hamburg.de}
  }

\begin{document}
\maketitle
\begin{abstract}
In this paper, we apply transformer-based Natural Language Generation (NLG) techniques to the problem of text simplification.  Currently, there are only a few German datasets available for text simplification, even fewer with larger and aligned documents, and not a single one with narrative texts. In this paper, we explore to which degree modern NLG techniques can be applied to 
German narrative text simplifications. 
We use Longformer attention and a pre-trained mBART model. Our findings indicate that the existing approaches for German are not able to solve the task properly. We conclude on a few directions for future research to address this problem.
\end{abstract}

\section{Introduction}
\subsection{Motivation}
With the rise of the internet, it has become convenient and often free to access an abundance of texts. However, not all people who have access can fully read and understand the texts, even though they speak the language that the text is written in. Most often this problem originates in the complex nature of the texts. Text simplification can help to overcome this barrier.

Narrative forms are one of the primary ways humans create meaning \citep{fellugaGeneralIntroductionNarratology2011}. Narrative texts, then, make an important contribution to how we describe and shape our environment. Simple language also contributes to involving as many people as possible in this process. Providing narrative texts in a 
Simple Language (\textit{Einfache Sprache}) version, enables a large audience to read them. So, we present the first approach for the automatic text simplification of German narrative texts.

\subsection{Related Works}
Automatic text simplification started in 2010 \cite{speciaTranslatingComplexSimplified2010} as statistical machine translation to the rule-based automatic text simplification task, using  a Portuguese corpus (4500 parallel sentences). 
The first German text simplification dataset was created by \citep{hanckeReadabilityClassificationGerman2012} to train a readability classifier. The dataset consisted of unaligned articles from one adult-targeting and one child-targeting journal, and
was later improved and enlarged by \cite{weissModelingReadabilityGerman2018}, which added unaligned transcripts from one adult-targeting and one child-targeting German TV news show. Similarly, \cite{aumillerKlexikonGermanDataset2022} published a German document-aligned dataset with lexicon articles for adults and for children.
The first sentence-aligned German simplification dataset was published in 2013 \citep{klaperBuildingGermanSimple2013} with 270 articles from five different websites, mainly of organizations that support people with disabilities.
In 2016 the first (rule-based) automatic text simplification system for German was released \cite{suterRulebasedAutomaticText2016}.
The first parallel corpus for data-driven automatic text simplification for German was introduced by  \cite{sauberliBenchmarkingDatadrivenAutomatic2020}. The corpus contains 3616 sentence pairs from news articles. They additionally were the first to use transformer models for German text simplification and found out that their corpus was not large enough to train them. 
\cite{battistiCorpusAutomaticReadability2020} collected a larger 
corpus with 378 text pairs, 
mostly from websites of governments, specialized institutions, and non-profit organizations.
\cite{riosNewDatasetEfficient2021} 
investigated the usage of an adapted mBART \citep{liuMultilingualDenoisingPretraining2020} version with Longformer attention \citep{beltagyLongformerLongDocumentTransformer2020} on Swiss newspaper articles. 
These results have been further improved with a sentence-based approach \cite{eblingAutomaticTextSimplification2022}. Most recently, the first detailed surveys about German text simplification have been released \citep{anschutzLanguageModelsGerman2023,stoddenDEPLAINGermanParallel2023,schomackerDataApproachesGerman2023}.

\section{Methods}

\paragraph{Longformer mBART}\label{sec:mbart}
Our goal was to train a document-level text generation model with a larger context ($>510$ input tokens; exceeding most transformer-based models). 
Longformer is the only model to our knowledge, which could extend the context on a pre-trained transformer model. 
We searched on \href{https://huggingface.co/}{huggingface.co} 
and filtered for text2text-generation models (8551), German (225), $>5 000$ downloads (30), and that they can perform a German-to-German translation task. This leaves only \href{https://huggingface.co/facebook/mbart-large-50}{\textit{facebook/mbart-large-50}} and \href{https://huggingface.co/facebook/mbart-large-cc25}{\textit{facebook/mbart-large-cc25}}, both introduced in \citep{liuMultilingualDenoisingPretraining2020}. We decided to take \textit{facebook/mbart-large-cc25} since it has been trained on fewer languages (25; in the CC25 dataset extracted from  \cite{wenzekCCNetExtractingHigh2020, conneauUnsupervisedCrosslingualRepresentation2020}) in comparison to \textit{facebook/mbart-large-50} (50). Because we reasoned that the greater the relative proportion of German in pre-training, the better. 
Our situation is very similar to \citep{riosNewDatasetEfficient2021}, so we base our methods on their approaches. mBART uses a specific input format consisting of the sentence and a language-tag. 
We additionally created two tags: \texttt{de\_OR} and \texttt{de\_SI} for Standard German and Simple German, respectively. 
Both of them are derived from the original German tag \texttt{de\_DE} (fifth-largest proportion in CC25) and only modified during our fine-tuning process.  
 Similar to \cite{riosNewDatasetEfficient2021}, we applied the Longformer conversion to the mBART model with a maximum input length of $1024$ and $512$ as the attention window size.

\paragraph{Domain Adaptation}
By using domain adaptation, we aim to enrich the vocabulary with previously unseen words and adapt the existing embeddings to the narrative text domain and the historical environment of the texts. 
After we created the longmbart-model we started the domain adaptation process. We downloaded all documents from TextGrid (\href{https://textgrid.de/web/guest/digitale-bibliothek}{textgrid.de}) in the category ``prose'' and  randomly sampled 60 documents. In a next step, we sentence-split the documents using spaCy (\href{https://spacy.io/}{spacy.io}), shuffled them and masked $15\%$ of the words. We used these masked and unmasked sentence-pairs for a single epoch training of the model. Both sides of the pair are tagged with the \texttt{de\_DE} tag. 
We used a learning rate of $3e\!\!-\!\!10$, an attention window size of $512$ during the conversion, a maximum input and output length of $70$, and a batch size of $8$.

\paragraph{Fine-Tuning}
We fine-tune our model on document-aligned German narrative texts, using 
three sources for Standard Language data: 1) \href{https://www.gutenberg.org/ebooks/}{gutenberg.org}, 2) \href{https://www.projekt-gutenberg.org/}{projekt-gutenberg.org}, and 3) \href{https://textgridrep.org/}{textgridrep.org}. We selected
\textit{Die Bremer Stadtmusikanten} (\texttt{mils-stadtmusikanten}), \textit{Der seltsame Fall von Dr Jekyll und Mr Hyde} (\texttt{eb-hyde}) and \textit{Der Schimmelreiter} (\texttt{pv-schimmelreiter}) as development set because their amount of words is close to the average amount of all samples in the fine-tuning dataset and they originate from different sources. For the same reasons, we selected \textit{Des Teufels rußiger Bruder} (\texttt{mils-bruder}), \textit{Der Graf von Monte Christo} (\texttt{eb-christo}) and \textit{Der Sandmann} (\texttt{pv-sandmann}) for testing. We used four sources for Simple Language texts: 1) \href{https://einfachebuecher.de/epages/95de2368-3ee3-4c50-b83e-c53e52d597ae.sf/de_DE/?ObjectPath=\%2FShops\%2F95de2368-3ee3-4c50-b83e-c53e52d597ae\%2FCategories\%2FB\%C3\%BCcher\%2F\%22B\%C3\%BCcher+in+Einfacher+Sprache+\%28A2\%2FB1\%29\%22\%2FKlassiker}{einfachebuecher.de} (\texttt{eb}), 2) \href{https://www.kindermannverlag.de/produkt-kategorie/weltliteratur-fuer-kinder-ab-6-jahre/}{kindermannverlag.de} (\texttt{kv}), and 3) \href{https://www.passanten-verlag.de/lesen/}{passanten-verlag.de} (\texttt{pv}), which consist of classic novels, as well as 4) the \textit{Märchen in Leichter Sprache} `Fairy Tales in Simple Language' from \href{https://www.ndr.de/fernsehen/barrierefreie_angebote/leichte_sprache/Maerchen-in-Leichter-Sprache,maerchenleichtesprache100.html}{ndr.de} (\texttt{mils}). 
The links to the Standard Language and Simple Language version can be found in Table \ref{tab:gnats-document-list} in the appendix. The \texttt{mils} samples include the complete text, while for the novels we use only the excerpts provided in the form of free reading samples (usually the first chapter of the text). We manually cut in the end of the Standard Language version to match the extent of the Simple Language version. 

\paragraph{Hyperparameter Setup}

Following \citep{riosNewDatasetEfficient2021}, we set the attention mode  \citep{beltagyLongformerLongDocumentTransformer2020} to sliding chunks (with overlap) and the attention window size to $512$.
Since our dataset is rather small, we turn gradient accumulation (\href{https://pytorch-lightning.readthedocs.io/en/latest/common/trainer.html?highlight=accumulate_grad_batches\#accumulate-grad-batches}{\texttt{accumulate\_grad\_batches}}) off. 
We use the Adam optimizer and optimize the learning rate with the PyTorch Lightning \href{https://pytorch-lightning.readthedocs.io/en/stable/api/pytorch_lightning.callbacks.LearningRateFinder.html}{\texttt{LearningRateFinder}} between $3e\!\!-\!\!20$ and $3e\!\!-\!\!1$. For Decoding we use beam search (size $=4$).

\section{Analysis and Evaluation}\label{sec:analysis-evaluation}

\subsection{Analysis}\label{sec:analysis} 

We manually compared the three generated output sequences of our test texts to the Standard Language version and the Simple Language version.  In summary, we found that 1) the model copies the input text to a very high degree without any modifications, 2) in cases where the model discarded parts of the inputs, it did not recognize the importance of the sequence, such as spelled-out antecedents for pronouns, and 3) it truncates rather randomly and without any semantic reason. 

For reasons of space, we only discuss \textit{Der Sandmann} in the appendix (section \ref{sec:sandmann-analysis}), on which we can show all the phenomena we want to discuss.

\subsection{Evaluation Measures}

\subsubsection{BERTscore, BLEU and ROUGE}
BERTscore \cite{zhangBERTScoreEvaluatingText2020} is currently the recommended \citep{alva-manchegoSuitabilityAutomaticEvaluation2021} way of comparing (generated) text simplification candidates and the (gold) references. It is a soft metric that yields high correlations with human judgments \citep{alva-manchegoSuitabilityAutomaticEvaluation2021}. We select \textit{google/mt5-base} as the underlying model, since it is the best performing model with Max Length $>1022$, German support, and a compatible transformers version \citep{zhangBERTScoreDefaultLayer2020}
(\textit{google/mt5-xl} and \textit{google/mt5-large} did not fit our hardware resources).
Following \citep{alva-manchegoSuitabilityAutomaticEvaluation2021} we use the BERTscore to determine the early stopping point during fine-tuning. We additionally employed 
two n-gram based approaches, 
BLEU \cite{papineniBleuMethodAutomatic2002} and ROUGE \citep{linROUGEPackageAutomatic2004}, because they are the most commonly used metrics for text generation.

\subsubsection{Entropy}
We use two flavors of Shannon entropy as a characterization, or measurement, of redundancy. In the basic implementation, we calculate the bag-of-words (BOW) entropy: $H(W)=\sum_{w\in W}\frac{count(w)}{n}\cdot-\log_2\left(\frac{count(w)}{n}\right)$, where $w$ is a word in the bag of words $W$, $count(w)$ is the frequency of $w$ in $W$, $n$ is the total size of $W$, and $H(W)$ is the text-level entropy.

In addition, we calculate the shortest-unique-prefix (SUP) entropy \citep{kontoyiannisyComplexityEntropyLiterary1997},
by calculating the length of the shortest prefix starting at $x_i$ that does not appear starting anywhere in the previous $i$ tokens $x_0, x_1,\ldots,x_{i-1}$. This prefix-length $l_i$ can be thought of as the length of the next unique substring after the past up to position ($i-1$) has been encoded. 
In other words, this metric measures the surprise value of a substring. The SUP entropy is calculated as: 
$\hat{H}_N=\left[\frac{1}{N}\sum_{i=1}^N \frac{l_i}{log(i+1)}\right]^{-1}$ with $N<M$, where $M$ is the largest possible index (= the sequence length $+1$). 
\citep{kontoyiannisyComplexityEntropyLiterary1997} do not elaborate on how $N$ should be chosen, so we set it to $\left\lfloor\frac{M}{2}\right\rfloor$. 


In both cases, we use spaCy to tokenize the generated output. We consider all tokens including punctuation marks and lowercase them.

\begin{table*}[!t]
    \centering\fontsize{9pt}{9pt}\selectfont
{
    \centering
    \begin{tabular}{c c c c c c c c c}
        \toprule
         Domain Adapt. & BERTscore$_{\text{F}1}$ & ROUGE-$l_{\text{F}1}$ & BLEU & SUP & BOW & Fine Tuning $\clubsuit$ & lr \\
        \midrule
         - & \textbf{0.682} & \textbf{0.127} & \textbf{1.43} & \textbf{1.000} & \textbf{6.685} & 0 & - \\
         - & 0.682 & 0.127 & 1.43 & 1.000 & 6.685 & 1 & 7.8e-20 \\
         - & 0.318 & 0 & 0 & 340.000 & 0.003 & 11 (100;10)  & 8.1e-07 \\
         50 texts & 0.301 & 0 & 0 & 123.666 & 0.038 & 1 & 3e-10 $\spadesuit$ \\
         50 texts & 0.301 & 0 & 0 & 123.666 & 0.038 & 0 & - \\
         100 texts & 0.301 & 0 & 0 & 123.666 & 0.038 & 0 & - \\
         100 texts & 0.301 & 0 & 0 & 123.666 & 0.038 & 1 & 3e-10 $\spadesuit$ \\
         100 texts & 0.298 & 0 & 0 & 49.666 & 0.0441 & 11 (100;10) & 3e-10 $\spadesuit$ \\
        \bottomrule
    \end{tabular}
    \vspace{-2mm}
    }
    \caption{Average performance of our models on the test texts. $\clubsuit:$ Best epochs with max epochs (and early stopping patience, if used, in parenthesis). $\spadesuit:$ The lr auto was unable to find an optimal learning rate; so we use a predefined value.}
    \label{tab:final-results}
    \vspace{-5mm}
\end{table*}

\subsection{Results and Discussion} \label{sec:discussion-future-work}

We analyze the model's performance via two kinds of metrics: similarity-based (BERTscore, BLEU and ROUGE) and entropy-based (SUP and BOW). Table \ref{tab:final-results} shows that the model without fine-tuning and domain adaptation performs the best both in terms of entropy and similarity. A single epoch of fine-tuning seems not to affect the models' performance, but fine-tuning it for $11$ epochs worsens it drastically. Similarly, domain adaption without and with $1$ epoch of fine-tuning drops below all non-domain-adapted models. Both domain adaptation set-ups ($50$ and $100$ documents) perform the same, so the number of domain adaptation documents seems to have no effect on the performance. Interestingly, with more fine-tuning ($11$ epochs) the SUP entropy is improved, while the BERTscore-similarity further drops. 

The model without domain adaptation and without fine-tuning performed the best and the more we trained the model, the more frequently individual text elements are repeated---first individual clauses, then words, and in the end only characters. These are results that no longer represent meaningful texts, let alone a high-quality text simplification. We did not manage to definitively conclude on reasons why both fine-tuning and domain adaptation do not outperform the pre-trained model. We assume that the main reason could be so-called catastrophic forgetting, which can occur in all scenarios where machine learning models are trained on a sequence of tasks and the accuracy on earlier tasks drops significantly. 
The model in our experiments was previously trained on inter-language translation (from one language to another) and we fine-tune it on intra-language translation (from one version of a language to another version of the same language). So, domain adaptation, being an intra-language task, differs from the original mBART task. The model's general text generation capability dropped after fine-tuning and domain adaptation. 
\cite{ramaseshAnatomyCatastrophicForgetting2021} demonstrate that forgetting is concentrated at the higher model layers and argue that it should be mitigated there. In their set-up, these layers change significantly and erase earlier task subspaces through sequential training of multiple tasks. All the mitigation methods they investigate  stabilize higher layer representations, but vary on whether they enforce more feature reuse, or store tasks in orthogonal subspaces.  There are several other possible reasons for this behavior and opportunities to improve the models' performance. In the following section, we give an outlook on possible ways of adjustment.

\section{Conclusion and Future Work} \label{sec:future-works}
In this paper, we apply existing transformer-based methods to generate text simplifications on document level. Furthermore, we investigate the usage of fine-tuning and domain adaptation.

Our work contributes to the field of automatic German text simplifications. This field is understudied, and future works that want to build on top of our and other previous works' findings could research the following areas:

\paragraph*{Catastrophic Forgetting}
\cite{yuAdaptSumLowResourceDomain2021} investigate catastrophic forgetting  and speculate that their second phase of pre-training results in some form of catastrophic forgetting for the pre-trained model, which could have hurt the adaptation performance. They recommend to use RecAdam \citep{chenRecallLearnFinetuning2020}, which mitigated the problem in their abstractive text summarization study.

\paragraph*{Repetition Problem}
\citep{fanHierarchicalNeuralStory2018} show that maximization-based approaches (such as beam search) tend to produce text that contains undesirable repetitions, and  stochastic methods tend to produce text that is semantically inconsistent with the given prefix. We use beam search in our approach and experience a significant increase of repetition during training. \cite{xuLearningBreakLoop2022} divide approaches for mitigating repetition into 1) training-based \citep{welleckNeuralTextGeneration2020,linStraightGradientLearning2021,xuLearningBreakLoop2022} and decoding-based  \citep{seeGetPointSummarization2017,fanHierarchicalNeuralStory2018,holtzmanCuriousCaseNeural2020} approaches. Recently, two new decoding approaches, Nucleus \citep{holtzmanCuriousCaseNeural2020} and Contrastive Search \citep{suContrastiveSearchWhat2022}, have shown promising results in terms of reducing repetition and improving the overall quality of generated text. Future work could apply these newer decoding methods to the task of document-level text simplification. However, although there is an increasing number of mitigating techniques, the causes of the repetition problem are still under-investigated. 

\paragraph*{Entropy}
Entropy metrics provide additional and very inexpensive guidance on the quality of generated text simplification. They can show very well to what degree the repetition problem is present in the text generation model. We encourage future research in similar tasks to measure entropy in their works.

\paragraph*{Masking Strategies}
We only use the commonly used token masking strategy for BART. \cite{lewisBARTDenoisingSequencetoSequence2019} describes other strategies, that can be used in the future as well.

\paragraph*{Controllability and Learning Strategies}
\cite[p. 1165--1168]{erdemNeuralNaturalLanguage2022} name a few resources where adding metadata, such as named entities or parts of speech, to the input can be used as an advanced learning strategy to improve results and offer more controllability over the output. We do not add any metadata and observed in Section \ref{sec:analysis-evaluation} that our model is not able to properly recognize named entities. Inserting corresponding metadata could potentially improve the performance in this regard.

\paragraph*{Unify Designations}
Designations of people are interchangeably used in Standard Language. A good example is the father in \textit{Der Sandmann}, who is mostly addressed as \textit{Vater} (``father'') but also as \textit{Papa} (``dad'') by his children and as \textit{Herr} (``master''), as in \textit{Herr des Hauses} (``man of the house''), by his house staff. All these words mean the same and are referring to the same person. Unifying them could help.

\newpage
\section*{Limitations}
The work we described in this paper investigates the automatic simplification of narrative documents in German. Our Methodology is focussed on document-level simplification and is only transferable to a limited extent to simplification that works on sentence-level or other linguistic levels. Additionally, our approach as well as the future research areas are generally applicable to document-level simplification in a broad variety of languages. The choice of quantitative evaluation can be applied to any text simplification task, with structural and linguistic limitations. The qualitative evaluation highly considers the narrative nature of our data, so it is transferable to the simplification of narrative texts in any form but hardly applicable to other text genres. 

The data we used is targeted towards different audiences, children and/or people with a lower literacy. Furthermore, some are written in easy language (\textit{Leichte Sprache}) and others in the broader category simple language (\textit{Einfache Sprache}). Future researchers are advised to carefully check the data sources and evaluate to which degree the data can be used for the intended purpose. Due to copyright restrictions, we are only able to provide public URLs to the data, and cannot provide the data directly.

\section*{Ethics Statement}
We state that our work complies with the ACL Ethics Policy.\footnote{\url{https://www.aclweb.org/portal/content/acl-code-ethics}} Our work investigates the automatic simplification of narrative documents in German. Providing simplified versions of texts positively contributes to the inclusion of people with cognitive disabilities and lower literacy into a growing number of aspects of society. Automatically generated simplifications offer a lower cost point compared to their human-made equivalents. On the one hand, this increases the number of people that can afford to read these text, on the other hand, it can endanger the future job prospects of human translators, which specialized in simplifying texts.

\section*{Acknowledgments}
We would like to thank the Norddeutscher Rundfunk (NDR) for allowing us to use the "Märchen in Leichter Sprache" \footnote{\url{https://www.ndr.de/fernsehen/barrierefreie_angebote/leichte_sprache/Maerchen-in-Leichter-Sprache,maerchenleichtesprache100.html}} and make them available to a scientific audience.

\bibliography{anthology,custom,2022MasterThesis}
\bibliographystyle{acl_natbib}

\newpage
\appendix
\section{Code}
Our code is available on GitHub:

\begin{description}
\item Pre-Processing of the Fine-Tuning Dataset based on Projekt Gutenberg, Gutenberg,  and PDF-Reading Samples Texts:\\ \href{https://github.com/tschomacker/aligned-narrative-documents}{github.com/tschomacker/aligned-narrative-documents}
\item Pre-Processing of the Domain Adapation based on Textgrid Texts:\\ \href{https://github.com/tschomacker/textgrid-domain-adaptation-dataset}{github.com/tschomacker/textgrid-domain-adaptation-dataset}
\item Machine Learning Architecture and Implementation:\\ \href{https://github.com/tschomacker/longmbart}{github.com/tschomacker/longmbart}
\end{description}

\newpage
\section{Analysis of \textit{Der Sandmann}}\label{sec:sandmann-analysis}
\begin{figure*}[!h]

  \raggedleft
\begin{fmpage}{0.95\linewidth}

\fontsize{8pt}{12pt}\selectfont
{
\begin{internallinenumbers}
    Ofel an LotharGewiß seid Ihr alle voll Unruhe, daß ich so lange - lange nicht geschrieben. Mutter zürnt wohl, und Clara mag glauben, ich lebe hier in Saus und Braus und vergesse mein holdes Engelsbild, so tief mir in Herz und Sinn eingeprägt, ganz und gar. - Dem ist aber nicht so; täglich und stündlich gedenke ich Eurer aller und in süßen Träumen geht meines holden Clärchens freundliche Gestalt vorüber und lächelt mich mit ihren hellen Augen so anmutig an, wie sie wohl pflegte, wenn ich zu Euch hineintrat. - Ach wie vermochte ich denn Euch zu schreiben, in der zerrissenen Stimmung des Geistes, die mir bisher alle Gedanken verstörte! - Etwas Entsetzliches ist in mein Leben getreten! - Dunkle Ahnungen eines gräßlichen mir drohenden Geschicks breiten sich wie schwarze Wolkenschatten über mich aus, undurchdringlich jedem freundlichen Sonnenstrahl. - Nun soll ich Dir sagen, was mir widerfuhr. Ich muß es, das sehe ich ein, aber nur es denkend, lacht es wie toll aus mir heraus. - Ach mein herzlieber Lothar! wie fange ich es denn an, Dich nur einigermaßen empfinden zu lassen, daß das, was mir vor einigen Tagen geschah, denn wirklich mein Leben so feindlich zerstören konnte! Wärst Du nur hier, so könntest Du selbst schauen; aber jetzt hältst Du mich gewiß für einen aberwitzigen Geisterseher. - Kurz und gut, das Entsetzliche, was mir geschah, dessen tödlichen Eindruck zu vermeiden ich mich vergebens bemühe, besteht in nichts anderm, als daß vor einigen Tagen, nämlich am 30. Oktober mittags um 12 Uhr, ein Wetterglashändler in meine Stube trat und mir seine Ware anbot. Ich kaufte nichts und drohte, ihn die Treppe herabzuwerfen, worauf er aber von selbst fortging.Du ahnest, daß nur ganz eigne, tief in mein Leben eingreifende Beziehungen diesem Vorfall Bedeutung geben können, ja, daß wohl die Person jenes unglückseligen Krämers gar feindlich auf mich wirken muß. So ist es in der Tat. Mit aller Kraft fasse ich mich zusammen, um ruhig und geduldig Dir aus meiner frühern Jugendzeit so viel zu erzählen, daß Deinem regen Sinn alles klar und deutlich in leuchtenden Bildern aufgehen wird. Indem ich anfangen will, höre ich Dich lachen und Clara sagen: 'Das sind ja rechte Kindereien!' - Lacht, ich bitte Euch, lacht mich recht herzlich aus! - ich bitt Euch sehr! - Aber Gott im Himmel! die Haare sträuben sich mir und es ist, als flehe ich Euch an, mich auszulachen, in wahnsinniger Verzweiflung, wie Franz Moor den Daniel. So ist es in der Tat. Mit aller Kraft fasse ich Euch aus meiner frühern Jugendzeit so viel zu erzählen, daß Deinem regen Sinn alles klar und deutlich in leuchtenden Bildern aufgehen wird. Indem ich anfangen will, höre ich Dich lachen und Clara sagen: 'Das sind ja rechte Kindereien!' - Lacht, ich bitte Euch, lacht mich recht herzlich aus! - es ist, als flehe ich Euch an, mich auszulachen, in wahnsinniger Verzweiflung, wie Franz Moor den Daniel. 
    \colorbox{Yellow!30}{So ist es in der Tat. So ist es in der Tat}
    . Mit aller Kraft fasse ich Euch aus meiner frühern Jugendzeit so viel zu erzählen, daß Deinem 
    \colorbox{Yellow!30}{unglückseligen Krämers gar feindlich auf mich wirken muß}
    , ja, daß wohl die Person jenes unglückseligen Krämers gar feindlich auf mich wirken muß, ja, daß wohl die Person jenes unglückseligen Krämers gar feindlich auf mich wirken muß. So ist in der Tat. So ist es in der Tat. - Nun fort zur Sache!Außer dem Mit aller dem 
    \colorbox{Yellow!30}{Mittagsessen, das alter Sitte gemäß schon um sieben Uhr aufgetragen wurde}
    , das alter Sitte gemäß schon um sieben Uhr aufgetragen wurde. 
    \colorbox{Yellow!70}
    {Er }
    \colorbox{Yellow!30}
    {mochte mit seinem Dienst}
    . Er mochte mit seinem Dienst viel beschäftigt sein. Nach dem Abendessen. Er mochte mit seinem Dienst viel beschäftigt sein. Nach dem Abendessen, das alter Sitte gemäß, das alter Sitte gemäß, das alter Sitte gemäß, das alter Sitte gemäß von uns um sieben Uhr aufgetragen. Nach dem Abendessen, daß er aber von selbst fortging, daß er aber von selbst fortging.
\end{internallinenumbers}
}
\end{fmpage}
    \caption{Generated output of our best performing model, with "Der Sandmann" by E. T. A. Hoffmann as input. We did not change the format of the besides adding highlights and line numbers. The yellow highlights point the reader towards text passages, which showcase our model's shortcomings, which we discuss in Section \ref{sec:sandmann-analysis}. For more information on the of the text, please refer to Table \ref{tab:gnats-document-list}.}
    \label{fig:sandmann-output}
\end{figure*}


The output of the model is shown in Figure \ref{fig:sandmann-output}. 
From line 1 to 18, the complete text is equivalent to the Standard Language input. After \textit{Franz Moor den Daniel} (line 19) the model inserts multiple passages that occur previously in the text, e.g. three times \textit{unglückseligen Krämers gar feindlich auf mich wirken muß} (``I can't help but think that the unfortunate grocer must have a hostile effect on me'') in line 23.

Furthermore, the model changes facts in the text, e.g. in line 26 supper is served at seven o'clock, while in the Standard Language version lunch is served at the same time. The reference Simple Language version from Passanten Verlag completely discards the facts about dinner and supper, boiling this passage down to a brief introduction of the father and mentioning that he was busy with his work and that he told fascinating stories to his kids. 
Another aspect of line 26 is that the model output does not mention the father. This is the first introduction of this character in the story, so the model discarded an important character from this text passage. Furthermore, although the model does not fully remove the father from the text, it only refers to him via pronouns: \textit{Er mochte mit seinem Dienst} (``He might be with his work'') (line 27) refers to the father by the pronoun \textit{Er} (``He''), despite the fact that the character was never introduced or referred to before. For a reader who has only access to the model's output, it is impossible to understand who \textit{Er} (``He'') is. A clean or complete removal of a character would show some simplification capability, even if it was an important character. In this case, it was an incomplete removal of an arguably important character.

Most of the repeated sentences do not contain information that is important to follow the story. In this respect, there is actually no need to transfer them into the simplification, let alone repeat them. Especially sentences like \textit{So ist es in der Tat} (``So it is indeed''), are only a linguistic emphasis and arguably add linguistic complexity without additional content. 
If we assume that repeated sentences are perceived as important by the model \footnote{Profound hypothesis on the causes of repetition are sparse. We base our conjecture on the results of \citet{xuLearningBreakLoop2022}, assuming a correlation between initial probability and repetition rate. If a text fragment occurs more often in the course of the document, it is more likely to be repeated. Therefore, we would say that \textit{unglückseligen Krämers} (``unfortunate grocer'') has a high initial probability for the model. In this respect, it is information that should be repeated more frequently in the text and can therefore be considered as important.}, the model correctly recognizes an importance only in one case, namely the first mention of the barometer seller Coppolla (one of the main characters in the story) and the narrator's fear of him in lines 26--28, \textit{unglückseligen Krämers gar feindlich auf mich wirken muß} (``unfortunate grocer must have a hostile effect on me'').

\begin{table*}[!t]
    \section{Additional Tables}
            
        \fontsize{8pt}{8pt}\selectfont
        {
        \rotatebox{90}{
                \begin{tabular}{l r l l}
                \toprule
                \textbf{Full Title} & Source-ID & Published & Source Texts (Standard and Simple)  \\
                \midrule
                    Die Abenteuer von Tom Sawyer & eb-sawyer & English (1876) &  \href{https://www.gutenberg.org/ebooks/30165}{gutenberg.org/ebooks/30165}  \\
                    & & &  \href{https://einfachebuecher.de/Die-Abenteuer-von-Tom-Sawyer/978-3-947185-33-7}{einfachebuecher.de/Die-Abenteuer-von-Tom-Sawyer/978-3-947185-33-7}\\
                 \rowcolor{gray!10}
                    Moby Dick & eb-moby & English (1851) &  \href{https://www.projekt-gutenberg.org/melville/mobydick/}{projekt-gutenberg.org/melville/mobydick/}     \\
                 \rowcolor{gray!10}
                    & & & \href{https://einfachebuecher.de/Moby-Dick/978-3-944668-86-4}{einfachebuecher.de/Moby-Dick/978-3-944668-86-4}  \\
                    Der Graf von Monte Christo &  eb-christo & French (1846) & \href{https://www.projekt-gutenberg.org/dumasalt/montchr1/}{projekt-gutenberg.org/dumasalt/montchr1/}   \\
                    & & & \href{https://einfachebuecher.de/Der-Graf-von-Monte-Christo/978-3-944668-53-6}{einfachebuecher.de/Der-Graf-von-Monte-Christo/978-3-944668-53-6}  \\
                 \rowcolor{gray!10}
                    Die Abenteuer von Huckleberry Finn &  eb-huckleberry  & English (1885) & \href{https://www.gutenberg.org/ebooks/64482}{gutenberg.org/ebooks/64482} \\ 
                 \rowcolor{gray!10}
                    & & & \href{https://einfachebuecher.de/Die-Abenteuer-von-Huckleberry-Finn/978-3-947185-34-4}{einfachebuecher.de/Die-Abenteuer-von-Huckleberry-Finn/978-3-947185-34-4} \\
                    Der seltsame Fall von Dr Jekyll und Mr Hyde & eb-hyde & English (1886) & \href{https://www.projekt-gutenberg.org/stevenso/jekyhyde}{projekt-gutenberg.org/stevenso/jekyhyde/} \\
                    & & & \href{https://einfachebuecher.de/Der-seltsame-Fall-von-Dr-Jekyll-und-Mr-Hyde/978-3-944668-54-3}{einfachebuecher.de/Der-seltsame-Fall-von-Dr-Jekyll-und-Mr-Hyde/978-3-944668-54-3}\\
                 \rowcolor{gray!10}
                    In 80 Tagen um die Welt & eb-welt & French (1873) & \href{https://www.projekt-gutenberg.org/verne/80tage/}{projekt-gutenberg.org/verne/80tage/}  \\
                 \rowcolor{gray!10}
                    & & & \href{https://einfachebuecher.de/In-80-Tagen-um-die-Welt/978-3-944668-32-1}{einfachebuecher.de/In-80-Tagen-um-die-Welt/978-3-944668-32-1}  \\
                    Aus Kinderzeiten (in \textit{Diesseits}) & eb-hesse & German (1907) & \href{https://www.gutenberg.org/ebooks/47818}{gutenberg.org/ebooks/47818}   \\
                    & & & \href{https://einfachebuecher.de/Erzaehlungen-von-Hermann-Hesse/978-3-944668-85-7}{einfachebuecher.de/Erzaehlungen-von-Hermann-Hesse/978-3-944668-85-7}  \\
                 \rowcolor{gray!10}
                    Sherlock Holmes. Das gesprenkelte Band & eb-band & English (1892) & \href{https://www.projekt-gutenberg.org/doyle/getupfte/chap002.html}{projekt-gutenberg.org/doyle/getupfte/chap002.html}  \\
                 \rowcolor{gray!10}
                    & & & \href{https://einfachebuecher.de/Sherlock-Holmes.-Das-gesprenkelte-Band/978-3-944668-36-9}{einfachebuecher.de/Sherlock-Holmes.-Das-gesprenkelte-Band/978-3-944668-36-9}  \\
                    Sherlock Holmes. Das Zeichen der Vier & eb-vier & English (1890) & \href{https://www.projekt-gutenberg.org/doyle/zeichen4/}{projekt-gutenberg.org/doyle/zeichen4/}  \\
                    & & & \href{https://einfachebuecher.de/Sherlock-Holmes.-Das-Zeichen-der-Vier/978-3-944668-39-0}{einfachebuecher.de/Sherlock-Holmes.-Das-Zeichen-der-Vier/978-3-944668-39-0}  \\
                 \rowcolor{gray!10}
                    20.000 Meilen unter dem Meer & eb-meer & French (1870) & \href{https://www.projekt-gutenberg.org/verne/zwanzig1/}{projekt-gutenberg.org/verne/zwanzig1/}  \\
                 \rowcolor{gray!10}
                    & & & \href{https://einfachebuecher.de/20.000-Meilen-unter-dem-Meer/978-3-947185-56-6}{einfachebuecher.de/20.000-Meilen-unter-dem-Meer/978-3-947185-56-6}  \\
                    Die Verwandlung & eb-verwandlung & German (1912) & \href{https://www.gutenberg.org/ebooks/22367}{gutenberg.org/ebooks/22367}    \\
                    & & & \href{https://einfachebuecher.de/Die-Verwandlung/978-3-947185-99-3}{einfachebuecher.de/Die-Verwandlung/978-3-947185-99-3}  \\
                 \rowcolor{gray!10}
                    Wolfsblut & pv-wolfsblut & English (1906) & \href{https://www.projekt-gutenberg.org/london/wolfsblu/}{projekt-gutenberg.org/london/wolfsblu/} \\
                 \rowcolor{gray!10}
                    & & & \href{https://www.passanten-verlag.de/lesen/\#wolfsblut}{passanten-verlag.de/lesen/\#wolfsblut}  \\
                    Der Schimmelreiter & pv-schimmelreiter & German (1888) & \href{https://www.projekt-gutenberg.org/storm/schimmel/}{projekt-gutenberg.org/storm/schimmel/} \\
                    & & & \href{https://www.passanten-verlag.de/lesen/\#schimmelreiter}{passanten-verlag.de/lesen/\#schimmelreiter}  \\
                 \rowcolor{gray!10}
                    Undine & pv-undine & French (1811) & \href{https://www.projekt-gutenberg.org/fouque/undine/}{projekt-gutenberg.org/fouque/undine/} \\
                 \rowcolor{gray!10}
                    & & & \href{https://www.passanten-verlag.de/lesen/\#undine}{passanten-verlag.de/lesen/\#undine}  \\
                    Hiob & pv-hiob & German (1930) & \href{https://www.projekt-gutenberg.org/roth/hiob/}{projekt-gutenberg.org/roth/hiob/}\\
                    & & & \href{https://www.passanten-verlag.de/lesen/\#hiob}{passanten-verlag.de/lesen/\#hiob}  \\
                 \rowcolor{gray!10}
                    Der Sandmann & pv-sandmann & German (1816) & \href{https://www.gutenberg.org/ebooks/6341}{gutenberg.org/ebooks/6341} \\
                 \rowcolor{gray!10}
                    & & & \href{https://www.passanten-verlag.de/lesen/\#sandmann}{passanten-verlag.de/lesen/\#sandmann}  \\
                    Weiße Nächte & pv-naechte & Russian (1848) & \href{https://www.projekt-gutenberg.org/dostojew/novellen/chap01.html}{projekt-gutenberg.org/dostojew/novellen/chap01.html} \\
                 
                    & & & \href{https://www.passanten-verlag.de/lesen/\#naechte}{passanten-verlag.de/lesen/\#naechte}  \\
                 \rowcolor{gray!10}
                    Der glückliche Prinz & pv-prinz & English (1888) & \href{https://www.projekt-gutenberg.org/wilde/maerche1/chap001.html}{projekt-gutenberg.org/wilde/maerche1/chap001.html} \\
                \rowcolor{gray!10}
                    & & & \href{https://www.passanten-verlag.de/lesen/\#prinz}{passanten-verlag.de/lesen/\#prinz3}  \\
                    Der Sandmann & kv-sandmann & German (1816) & \href{https://www.gutenberg.org/ebooks/6341}{gutenberg.org/ebooks/6341} \\
                    & & & \href{https://www.kindermannverlag.de/produkt/der-sandmann/}{kindermannverlag.de/produkt/der-sandmann/}  \\
                 \rowcolor{gray!10}
                    Der Schimmelreiter & kv-schimmelreiter & German (1888)  & \href{https://www.projekt-gutenberg.org/storm/schimmel/}{projekt-gutenberg.org/storm/schimmel/} \\
                 \rowcolor{gray!10}
                    & & & \href{https://www.kindermannverlag.de/produkt/der-schimmelreiter/}{kindermannverlag.de/produkt/der-schimmelreiter/}  \\
                    Kinder- und Hausmärchen - Brüder Grimm & All mils-documents  & German (1858) & \href{https://www.projekt-gutenberg.org/grimm/khmaerch/}{projekt-gutenberg.org/grimm/khmaerch/} \\
                     & & & \href{https://www.ndr.de/fernsehen/barrierefreie_angebote/leichte_sprache/Maerchen-in-Leichter-Sprache,maerchenleichtesprache100.html}{ndr.de/fernsehen/barrierefreie\_angebote/leichte\_sprache/Maerchen-in-Leichter-Sprache,}  \\
                     & & & \hspace{3mm}\href{https://www.ndr.de/fernsehen/barrierefreie_angebote/leichte_sprache/Maerchen-in-Leichter-Sprache,maerchenleichtesprache100.html}{maerchenleichtesprache100.html}  \\
                    \bottomrule
            \end{tabular}
            }
            \caption{All documents in our corpus from \url{einfachebuecher.de} (\texttt{eb}) which are classified as ``Klassiker'' (classic novel) (snapshot from 07/14/2022), and Passanten Verlag (\texttt{pv}) (snapshot from 07/14/2022), Kindermann Verlag (\texttt{kv}) (snapshot from 07/14/2022) and \textit{Märchen in Leichter Sprache} (\texttt{mils}) (snapshot from 07/14/2022).}
            \label{tab:gnats-document-list}
}
        \end{table*}

\end{document}